\documentclass{article}

\usepackage[letterpaper]{geometry}
\usepackage{spconf,amsmath,epsfig}
\usepackage{enumitem}

\usepackage{fancyhdr}
\usepackage{color}

\def\onedot{.\ }

\def\etal{\emph{et al.\ }}

\begin{document}\sloppy

\def\x{{\mathbf x}}
\def\L{{\cal L}}

\title{Joint Background Reconstruction and Foreground Segmentation via A Two-Stage Convolutional Neural Network}
%
\name{Xu Zhao$^{1,2}$, Yingying Chen$^{1,2}$, Ming Tang$^{{*},1,2}$ and Jinqiao Wang$^{1,2}$\thanks{This work was supported by Natural Science Foundation of China. Grant No. 61375035.}
\thanks{{*}Corresponding author.}}
\address{
$^{1}$National Laboratory of Pattern Recognition, Institute of Automation\\
    Chinese Academy of Sciences, Beijing, China, 100190\\
$^{2}$University of Chinese Academy of Sciences, Beijing, China, 100190\\
\{xu.zhao, yingying.chen, tangm, jqwang\}@nlpr.ia.ac.cn}

\maketitle

\begin{abstract}
Foreground segmentation in video sequences is a classic topic in computer vision.
Due to the lack of semantic and prior knowledge, it is difficult for existing methods
to deal with sophisticated scenes well.
Therefore, in this paper, we propose an end-to-end two-stage deep convolutional
neural network (CNN) framework for foreground
segmentation in video sequences. In the first stage, a convolutional encoder-decoder
sub-network is employed to reconstruct the background images and encode rich prior
knowledge of background scenes.
In the second stage, the reconstructed background and current frame are input into a multi-channel
fully-convolutional sub-network (MCFCN) for accurate foreground segmentation.
In the two-stage CNN, the reconstruction loss and segmentation loss are
jointly optimized. The background images and foreground objects are output
simultaneously in an end-to-end way.
Moreover, by incorporating the prior semantic knowledge of foreground and background
in the pre-training process, our method could
restrain the background noise and keep the integrity of foreground objects at the same time.
Experiments on CDNet 2014 show that our method outperforms the state-of-the-art by 4.9\%.

\end{abstract}
\section{Introduction}
\label{sec:intro}
Foreground segmentation in video sequences is classic topic in computer vision.
It is the fundamental step for many high-level computer vision tasks, such as activity recognition, object tracking and automated anomaly detection in video surveillance.
In the last two decades, many algorithms have been proposed to solve this problem \cite{st2015subsense, st2015self, barnich2011vibe, stauffer1999adaptive, Elgammal2000, Allebosch2016, wang2014static, chen2015learning, Chen2016A}.

Statistically modeling background is a prevalent strategy to extract foreground for its robustness and efficiency. Some typical methods, such as GMM \cite{stauffer1999adaptive}, KDE \cite{Elgammal2000}, and ViBe \cite{barnich2011vibe}
assume the independence among pixels and model the variation of every pixel over time. Several other works \cite{Allebosch2016, st2015subsense, wang2014static} attempt to employ more discriminative hand-crafted features. Nevertheless, all existing methods perform poorly in scenes where the night light, heavy shadows, or camouflaged foreground are involved, because they all do not consider the semantic explanation of the scenes. And usually, post-processing has to be employed by them to smooth the foreground edges and reduce the noises on the foreground.

In general, the existing background modeling methods are considered unsupervised.
The fact, however, is that there still exist many parameters in them to control the initialization, classification, and update rate of background model, etc., and unfortunately, all these parameters have to be set to adapt to different characteristics of videos.
In order to achieve high performance on the whole benchmark, a variety of experiential rules for setting the parameters have to be designed according to characteristics of the whole benchmark. This implies that the background model is actually \emph{tuned with the test data}.
Therefore, it could be argued that the existing background modeling methods might not be considered as unsupervised.
It would be more reasonable to exploit some labeled data to train the model, and others to test it.

\begin{figure*}
\centering
    \includegraphics[height=0.20\textheight]{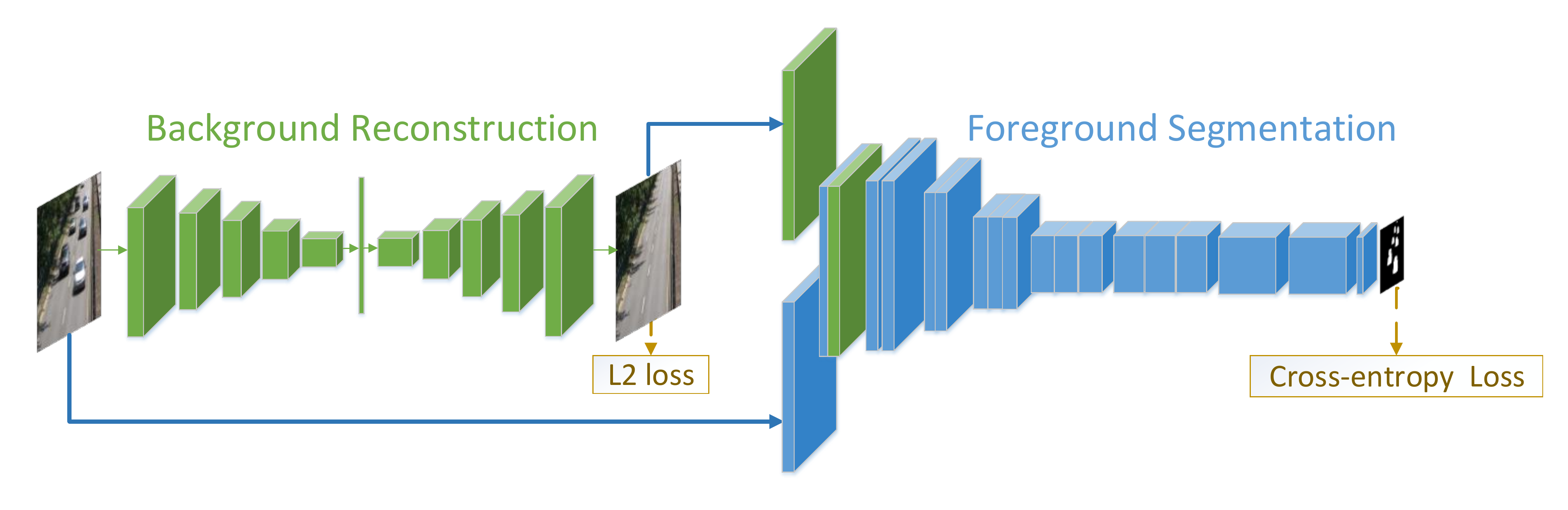}

   \caption{An overview of the two-stage deep neural network for foreground segmentation. The current frame is input into the encoder-decoder network to reconstruct the background. Then the reconstructed background image is concentrated to the current frame and fed into the following fully convolutional network to segment the foreground.}
\label{fig:overview}
\end{figure*}

Therefore, in this paper, we propose a supervised end-to-end
two-stage deep convolutional neural network to segment the foreground regions,
utilizing semantic information learned by the network.
An overview of the network is shown in Fig.\ \ref{fig:overview}.
It consists of a background reconstruction stage and a foreground segmentation stage.
Our network is designed based on the deep fully-convolutional network (FCN).
FCNs can extract the semantic features of the input. And through supervised learning, it shows good performance
in pixel-wise labeling tasks, such as semantic segmentation \cite{long2015fully}. However, the original
FCN takes a single image as input, so that it lacks the analysis on the foreground motion in its segmentation results,
which is necessary to distinguish moving objects from non-moving objects in the foreground segmentation task.
To combat this, we restore the background image first, and then concatenate it with the input image as multi-channel input for segmentation network. Consequently, the motion information of the foreground is naturally implied in the image pair of the input image and its reconstructed background.

Specifically, in the background reconstruction stage, we
adopt a deep convolutional encoder-decoder network,
which establishes the background model according to training
images and outputs the refined reconstructed background.
Such a background model encodes rich prior knowledge of background scenes and
can handle various background changes and generate clean background images.
In the segmentation stage, we adopt a multi-channel fully-convolutional network (MCFCN), which is a variant
of FCN. It processes a multi-channel input, which is a six-channel concatenation of the reconstructed background and the current image. The image pair of reconstructed background and current image naturally contains motion information of the foreground. Therefore, besides utilizing the semantic information, the MCFCN can distinguish the moving objects more accurately than the FCN that has single-image input.
The two sub-networks are jointly trained for the final optimization, so that two tasks of background reconstruction and foreground
segmentation sub-networks benefit from each other in the process of iteration.
\label{sec:overview}

The contribution of this paper is three-fold.
First, we propose a new two-stage neural network architecture for foreground segmentation,
which encodes rich background knowledge and combines semantic features with
motion information.
Second, with the reconstructed background and current frame as input together,
we adopt a multi-channel fully-convolutional network to
segment the foreground objects,
Third, the end-to-end network is jointly optimized with a multi-task loss
including reconstruction loss and segmentation loss.
The performances of the two sub-networks are mutually promoted by joint optimization.

\section{Related work}
\label{sec:Related}
\textbf{Background Reconstruction.}
Traditional foreground detection approaches often reconstruct the background by PCA \cite{oliver1999bayesian} or
RPCA \cite{candes2011robust,Rodriguez2016}.
Oliver \etal \cite{oliver1999bayesian} proposed a Principal Component Analysis
(PCA) based method that builds a vectorization representation with a set of images, and decomposes the representation
via PCA to restore the background.
In the last decade, Robust PCA (RPCA) based methods are proposed to decompose the foreground and background. Candes \etal \cite{candes2011robust} proposed
a convex optimization to address the RPCA-PCP problem. Recently, Rodriguez and Wohlberg \cite{Rodriguez2016}
proposed an incremental PCP algorithm that consumes low memory and calculation.

\textbf{Convolutional Autoencoder.} Denoising autoencoders \cite{vincent2008extracting} take corrupted image as input and then output the original image. But the original denoising autoencoders \cite{vincent2008extracting} can only undo damages that are localized and low-level. In \cite{pathak2016context} Pathak \etal presented Context Encoders,
a convolutional encoder-decoder network that can generate
the contents of an arbitrary missing image region conditioned on its surroundings.
In Context Encoders, the missing image regions can be regarded as noises. Thus, Context
Encoders share the same characteristics with denoising autoencoders.
Regarding the foreground as noises, we use a similar convolutional encoder-decoder framework
for background reconstruction as Context Encoders and get a satisfactory reconstruction result.

\textbf{Semantic Segmentation.} FCN \cite{long2015fully} has shown good performance in semantic segmentation.
Based on FCN, Deeplab \cite{chen2016deeplab} architecture uses atrous convolutions to produce a more dense feature map and uses
CRF \cite{koltun2011efficient} to get a better localization especially
along the edge of objects. We adopt almost the same network architecture as the VGG-16 based DeepLab-LargeFOV
in the foreground segmentation sub-network, except some modifications like that
the first layer is changed to take a 6-channel input, which is the concatenation of the reconstructed background image and the current input frame.


For paired-image inputs,
siamese networks \cite{Chopra2005Learning} has been widely used.
Siamese networks is a two-stream network architecture that has two same
branches.
Without using siamese networks, Dosovitskiy \etal \cite{dosovitskiy2015flownet} and Khan \etal \cite{khan2016weakly} adopted one-stream multi-channel networks to process two input images in optical flow prediction and weakly supervised change detection.
The one-stream multi-channel network is more efficient than the siamese networks.
In the second stage, we adopt the one-stream multi-channel network architecture to locate the foreground regions in the image, inputting the current frame image and background image.

\section{Overview}

An overview of the proposed two-stage network is shown in Fig.\ \ref{fig:overview}. The first-stage sub-network is a convolutional encoder-decoder network for encoding rich background knowledge and background reconstruction. Next, the second-stage sub-network is an MCFCN to segment the foreground objects by inputting the current frame and its reconstructed background image. Finally, the whole network is jointly optimized with a multi-task loss.

\subsection{Encoder-decoder Network for Background Reconstruction}
Inspired by the network in \cite{pathak2016context} used for inpainting, we adopt an encoder-decoder network to reconstruct the background image.
The encoder contains a set of convolutions, and represents the input image as a latent feature vector.
The decoder restores the background image from the feature vector.
Generally, in the encoder-decoder networks \cite{hinton2006reducing,  pathak2016context,vincent2008extracting},
the latent feature vector distils the most
useful information from the input image. In our encoder-decoder network,
the vector represents
the context information and encodes the rich knowledge of background scenes.
Then the decoder restores the clean background image from the latent feature vector.

Our encoder-decoder network is designed following the
pattern as DCGAN \cite{radford2015unsupervised} and Context Encoders \cite{pathak2016context}.
The layers of the encoder-decoder network are all convolutional layers.
In the encoder part, the number of nodes in each layer is reduced layer by layer.
As in \cite{radford2015unsupervised} and \cite{pathak2016context}, we use strided convolution layers with the stride of 2 instead of using spatial pooling layers to downsample the layer's input.
We make the convolution layers learn a pooling operation without limited
to ``max'' or ``mean'' operation \cite{radford2015unsupervised}, which allows
the network to learn customized downsampling function.

We choose the L2 loss as the reconstruction loss, and the loss function of the encoder-decoder network can be expressed as follows:
\begin{equation}
  L_{rec}=\sum_{i,j,c}\|B_{ijc}-B^{*}_{ijc}\|_{2}, \label{ereg}
\end{equation}
where $B_{ijc}$ is the value of the reconstruction background image on the i-th row, the j-th colume and c-th color channel.
And $B^{*}$ is the ground truth background image.

The ground truth background images for each training frames are manually generated according to their label maps.
After training,
the encoder-decoder network can separate the background from the input image and restore a clean background image.
This is a good stepping-stone for the following segmentation step.

\subsection{MCFCN for Foreground Segmentation}
The second stage of our network is a multi-channel fully-convolutional network (MCFCN).
Its input is the concatenation of the reconstructed background image and the current frame.
The network
can learn semantic knowledge of the foreground and background.
Therefore, it could handle various changes better, like the night light, shadows and camouflaged foreground objects.
And by inputting the image pair of the background image and current frame, it can easily catch motion information, thus inferring the moving object easily.
Besides, according to our experiment results,
the network segments rather refined foregrounds without any post processing such as CRF when adding the background image as input.

As mentioned in \cite{dosovitskiy2015flownet, khan2016weakly} and proven by our initial experiment,
a multi-channel network has comparable
performance to the traditionally siamese network \cite{Chopra2005Learning} for paired images.
However, the siamese network consumes as almost twice the memory as an MCFCN.
Therefore,
we take the 6-channel input by concatenating the reconstructed background image to the current frame, and then adopt
MCFCN to segment the foreground regions from the input.

The MCFCN derives from the VGG-16 based DeepLab-LargeFOV network \cite{chen2016deeplab},
and we make two modifications.
The one is
that original first layer with 3-channel input is substituted with a new layer taking 6-channel input.
The other is to change the arous sampling rate in the ``fc6'' layer from 12 to 6.
This leads to a smaller receptive field, which brings benefit to segment small objects.

For an input image, given its ground truth of
foreground segmentation map
$L$
and the output probability map $P$
from the softmax layer of the network, the segmentation loss is calculated as,
\begin{equation}
  L_{seg}=\sum_{i,j,k}\log P_{i,j,k}\boldsymbol{1}[k=L_{i,j}]\boldsymbol{1}[L_{i,j}\ne l^{ign}],
 \label{eseg}
\end{equation}
where $\boldsymbol 1$ is the indicator function. $P_{i,j,k}$ is the probability for class $k$
of the element on the $i$-th row and the $j$-th column. And $L_{i,j}\in\{0,1,-1\}$ is the ground truth label
, for which $1$ stands for the foreground, $0$ stands for the background and $-1$ stands for the ignored regions ($l^{ign} = -1$).

\subsection{Joint Optimization}
\textbf{Multi-task Loss.}
Previous work on object detection \cite{Girshick2015Fast} indicates that multi-task learning usually leads to an improvement on performance in tasks that relate to each other.
In our network, the two stages can be integrated into an end-to-end architecture and trained jointly.
Combining Eq. \ref{ereg} and Eq. \ref{eseg}, the multi-task loss can be represented as:
\begin{equation}
  L = L_{rec} + \lambda \times L_{seg},
  \label{ejoint}
\end{equation}
where $\lambda$ is a weight parameter.

In this way, the loss computed at the segmentation stage can be back-propagated to the encoder-decoder network. The two losses ($L_{rec}$ and $L_{seg}$) lead to a more refined background reconstruction and segmentation maps.

\textbf{Training Details.} We train the two-stage network in three steps.
In the first step, we train the encoder-decoder sub-network using single loss $L_{reg}$.
In the second step, we fix the parameters of the encoder-decoder
sub-network, and train the MCFCN using single loss $L_{seg}$.
Finally, we train the two sub-networks jointly
with the parameters of two pre-trained sub-networks.
The three-step training assures a faster convergence to
the optimal value than directly optimizing the entire network in one single step.

The three steps all adopt the mini-batch stochastic gradient descent (SGD) method to optimize the network.
The mini-batch sizes for the three steps are 4, 2 and 1 respectively. And the learning rate is set to $10^{-4}$, $10^{-3}$ and $10^{-5}$ respectively.
The first step trains the sub-network with 20k iterations. The second and the third step take 6000 and 3000 iterations respectively.
The MCFCN is initialized with the weights of pre-trained DeepLab model \cite{chen2016deeplab}.
And the other layers of the whole network is initialized from a zero-centered Normal distribution with standard deviation 0.01.
\begin{figure}[t]
\begin{center}
  \includegraphics[width=0.43\textwidth]{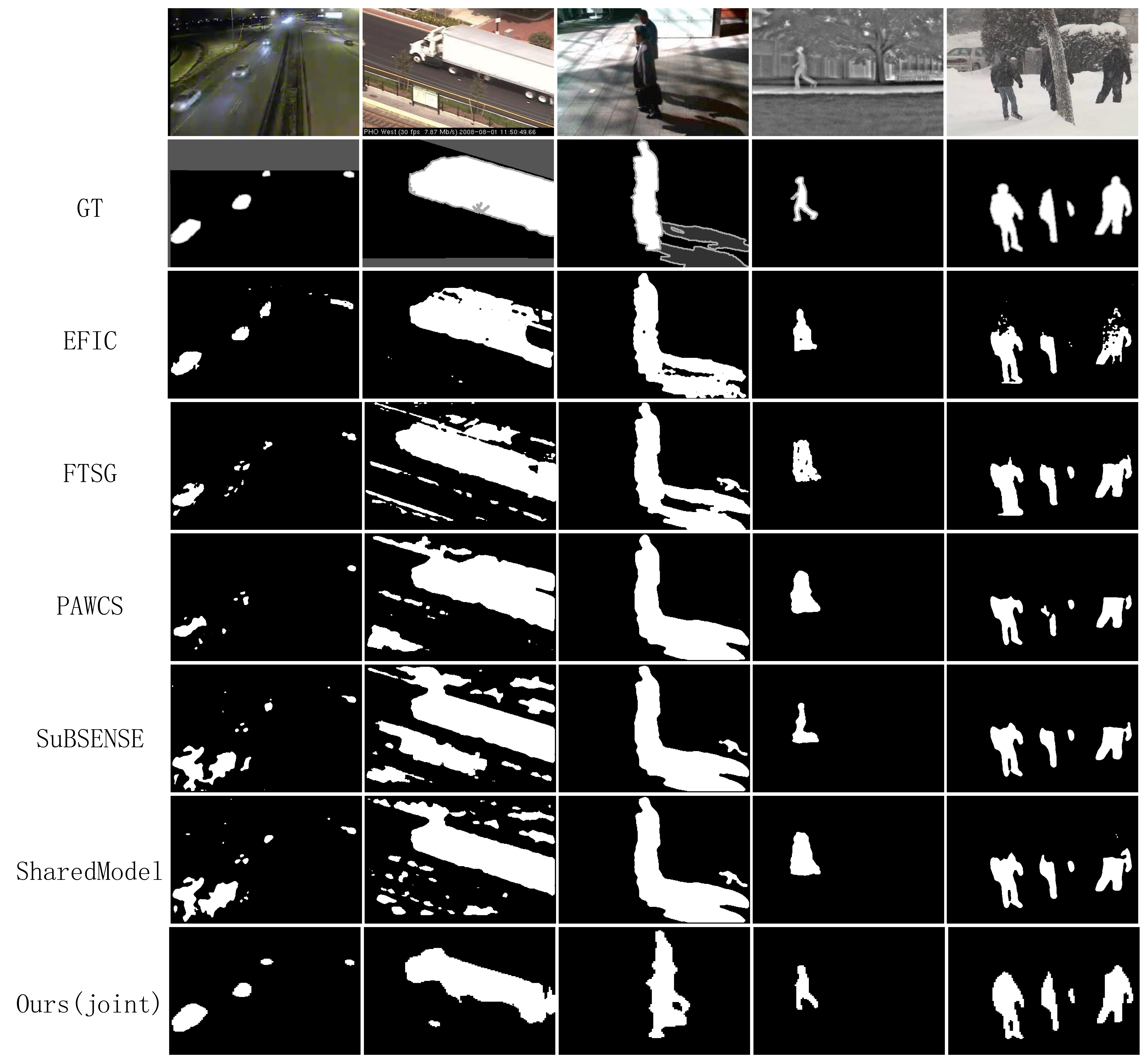}
\end{center}
   \caption{Visual comparison of foreground segmentation results. From top to bottom, the sequences are ``fluidHighway" from \emph{Night Videos},
   ``boulevard" from \emph{Camera Jitter},
``peopleInShade" from \emph{shadow}, ``park" from \emph{Thermal} and ``skating" from \emph{Bad Weather}.}
\label{fig:com}
\end{figure}
\subsection{Implementation Details}
\textbf{Framework Details.} We use the modified CAFFE \cite{jia2014caffe} code released with DeepLab v2 \cite{CP2015Semantic,chen2016deeplab} to implement the whole neural network. The input image ($I$) is resized to $128 \times 128$ ($I_1$) and $961\times 961$ ($I_2$), and $I_1$ is fed into the first sub-network to get the reconstructed background image ($B$) with the size of $128\times 128$. Then $B$ is resized to size $961 \times 961$ and fed into the second sub-network with $I_2$ together. The resize operation between the two sub-networks is implemented by a neural network layer that is initialized with bilinear interpolation's weights. After initialization, this layer's parameters do not update in the training process. Our initial experiment shows the parameter $\lambda$ in Eq.\ \ref{ejoint} has little impact on the performance. So,
$\lambda$ is set to 1 in this paper. On a computer with Intel i7 CPU with 4.00GHz and a single NVIDIA GTX Titan X GPU, our algorithm runs at five fps on test videos.

\section{Experiments}\label{sec:exp}
\subsection{Dataset}

We conduct experiments on the public ChangeDetection
benchmark 2014 (CDNet 2014) \cite{wang2014cdnet} to evaluate the performance of our algorithm.
CDNet 2014 dataset has 53 video sequences belonging to 11 diverse categories. All
video sequences are captured in real scenes.
Since ground truth is required in the training step of our network, we
build up the training and testing data set by dividing the frames having ground truth.
Specifically, suppose there are $n$ frames having ground truth,
numbered as $1, \dots, n$, the frames numbered as $[1,\lfloor\frac{n}{2}\rfloor]$ are used as training samples and
$[\lfloor\frac{n}{2}\rfloor+1,n]$ are used for testing.
We select the F-Measure to evaluate the performance, the same as CDNet 2014.
\begin{figure}
\begin{center}
  \includegraphics[width=0.39\textwidth]{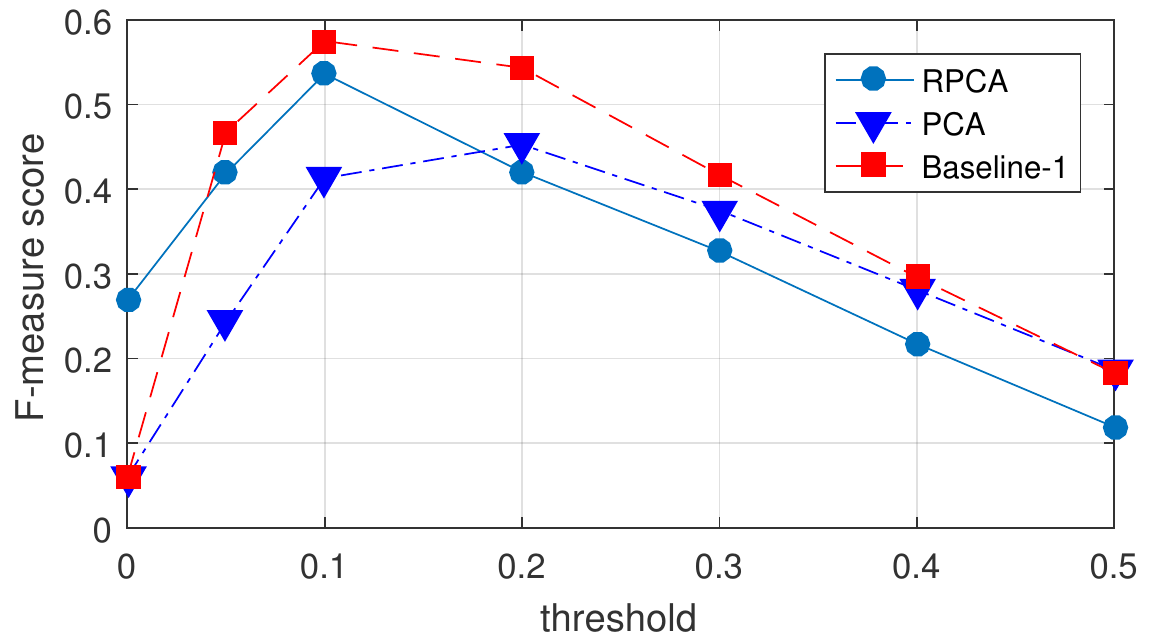}
\end{center}
   \caption{F-measure scores of three methods (PCA, RPCA, Baseline-1) varing with different thresholds.}
\label{fig:short}
\end{figure}


%

We train our network on each video's training frames separately.
For some videos, all their training frames have no foreground regions.
In this case, we randomly paste 1 or 2 objects as the foreground, which are cut from the object regions of PASCAL VOC 2012 \cite{Everingham2015} .

\subsection{Experiments on Background Reconstruction}


\begin{table*}
\small
  \begin{center}
    \caption{F-measures for subset of ChangeDetection benchmark 2014 \cite{wang2014cdnet}.PTZ: Pan/Tilt/Zoom; BW: Bad Weather; Ba: Baseline; CJ: Camera Jitter; DB: Dynamic Background; IOM: Intermittent Object Motion; LF: Low Framerate; NV: Night Videos; Sh: Shadow; Th: Thermal; Tu: Turbulence; Overall is the average F-measure of 11 categories. C-EFIC: \cite{Allebosch2016}, FTSG: \cite{wang2014static}, PAWCS: \cite{st2015self}, SuBSENSE: \cite{st2015subsense}, SharedModel: \cite{chen2015learning}, GMM: \cite{stauffer1999adaptive}, KDE: \cite{Elgammal2000},ViBe \cite{barnich2011vibe}.}
     \begin{tabular}{|p{2.3cm}p{0.8cm}p{0.8cm}p{0.8cm}p{0.8cm}p{0.8cm}p{0.8cm}p{0.8cm}p{0.8cm}p{0.8cm}p{0.8cm}p{0.8cm}|p{0.85cm}|}
         \hline 
        Approach &PTZ &BW &Ba &CJ &DB &IOM &LF &NV&Sh&Th&Tu&Overall \\ 
        \hline \noalign{\smallskip}\hline
        KDE &$0.0512 $&$0.7522 $&$0.9101 $&$0.5748 $&$0.6226 $&$0.3911 $&$0.5097 $&$0.4057 $&$0.8157 $&$0.7573 $&$0.5004 $&$0.5719 $\\
        GMM &$0.1937 $&$0.7819 $&$0.7845 $&$0.5920 $&$0.6650 $&$0.4841 $&$0.5008 $&$0.4003 $&$0.8047 $&$0.6377 $&$0.5364 $&$0.5801 $\\
        ViBe &$0.0666 $&$0.7734 $&$0.8793 $&$0.4476 $&$0.6357 $&$0.4726 $&$0.3286 $&$0.4019 $&$0.8241 $&$0.5512 $&$0.6058 $&$0.5443 $\\
        C-EFIC &${\color{red}0.6140}$&$0.7731$&$0.9382$&$0.8216$&$0.5531$&$0.5868$&$0.6984$&$0.6502$&$0.8808$&$0.8097$&$0.7230$&$0.7317$\\
        FTSG &$0.3923$&$\color{red}{0.8672}$&$0.9321$&$0.7066$&$\color{blue}{0.8750}$&$\color{green}{0.8423}$&$0.7133$&$0.5347$&
        $0.8876$&$0.7791$&$\color{green}0.7922$&$0.7566$\\
        PAWCS &$\color{blue}{0.5728}$&$0.7937$&$0.9384$&$0.7880$&$\color{red}{0.8995}$&$0.7756$&$\color{green}{0.7531}$&$0.4193$
        &$0.9016$&$0.7975$&$0.7583$&$\color{green}0.7634$\\
        SuBSENSE &$0.4165 $&$\color{blue}{0.8665} $&$\color{green}{0.9527} $&$0.7995 $&$0.8043 $&$0.7259 $&$0.6859 $&$0.5127 $&$\color{green}0.9037 $&$0.7820 $&$\color{red}{0.8600}$&$0.7554 $\\
        SharedModel &$0.4567$&$0.8072$&$0.9558$&$0.8034$&$\color{green}{0.8047}$&$0.6834$&$\color{blue}{0.7936}$&$0.4794$&$
        0.8907$&$0.7889$&$\color{blue}0.8493$&$0.7557$\\
        \hline\noalign{\smallskip}\hline

        Baseline-1 &$0.2815$&$ 0.7505$&$ 0.8681$&$ 0.7587$&$ 0.2369$&$ 0.6289$&$ 0.5780$&$0.3961$&$0.7450$&$0.7651$&$0.2993$&$0.5731$\\
        Baseline-2 &$0.5150$&$ 0.7228 $&$0.9048 $&$\color{blue}{0.8706} $&$	0.6796$&$ 	0.7779 $&$	0.7267 $&$	\color{red}{0.8042} $&$	0.8954 $&$	\color{green}0.8373 $&$	0.4242 $&$	0.7417$\\

        Ours &$0.4493 $&$0.8004 $&$\color{blue}{0.9630} $&$\color{green}{0.8699} $&$0.7405 $&$\color{blue}{0.8734} $&$\color{red}{0.8075} $&$\color{green}{0.6851} $&$\color{blue}{0.9216} $&$\color{blue}{0.8536} $&$0.6929 $&$\color{blue}0.7870 $\\
        Ours(joint)
        &$\color{green}{0.5168}$&$\color{green}{0.8550}$&$\color{red}{0.9680}$&$\color{red}{0.8988}$&$0.7716$&$\color{red}{0.9066}
        $&$0.7491$&$\color{blue}{0.7695}$&$\color{red}{0.9286}$&$\color{red}{0.8586}$&$0.7143$&$\color{red}{0.8124}$\\
   %
        \hline
      \end{tabular}
    \label{tab:cdnet}
  \end{center}
\end{table*}


To verify the performance of background reconstruction, we design the first baseline (named Baseline-1) by appending a threshold classifier to the reconstruction network.


We compare the Baseline-1's foreground segmentation results with original PCA-based method (PCA for short) \cite{oliver1999bayesian} and an incremental PCP-based RPCA method (RPCA for short) \cite{Rodriguez2016} on CDNet 2014.
The three methods all use the threshold-based classifier to
segment the foreground regions.
We change the threshold from 0 to 0.5 to find the best value for each method. Fig.\ \ref{fig:short} shows how the
F-measure varies with the threshold. It is clear that our method has
better performance than other methods.
The reason is that the background reconstruction of our method is more accurate than others.

In spite of the good performance on background reconstruction, a threshold-based classifier only achieves an F-measure less than 0.6. This
classifier can not handle
cases like the night light, shadows and camouflaged foreground well.
We need a better classifier to segment the foreground regions with the reconstructed background image and input frame.
\subsection{Experiments on Foreground Segmentation}


To evaluate how much the reconstructed background image benefits the segmentation result,
of the foreground segmentation sub-network,
we design the second baseline (Baseline-2), which segments the foreground using a 3-channel FCN sharing the same structure with
our multi-channel FCN at the second stage. It takes the current image as input and produces the foreground
segmentation result.
We train the Baseline-2 in the same way as our foreground segmentation sub-network.
We test both our two-stage network and Baseline-2 on the CDNet 2014 dataset \cite{wang2014cdnet} and compare their performance.

In Tabel \ref{tab:cdnet}, it is clear that the two-stage network performs better in most scenes
than Baseline-2.
Thus, we conclude that the segmentation network can infer the foreground regions more accurately by adding background image as input.
\subsection{Evaluation on ChangeDetection benchmark}
We compare the results of our two-stage network with the state-of-the-art methods and several classic methods on CDNet 2014
in Table \ref{tab:cdnet}.
We download the foreground segmentation results of some method \cite{Allebosch2016, wang2014static, st2015self, st2015subsense,chen2015learning,stauffer1999adaptive,Elgammal2000,barnich2011vibe} from CDNet\footnote{http://changedetection.net/} and calculate the F-measure on our testing data.
Besides, the result of ViBe is obtained with BGSLibrary \cite{Sobral2013BGSLibrary}.
In Table \ref{tab:cdnet}, our method with joint optimization achieves the best performance in 5 out of the 11 categories.
The two-stage network improves the state-of-the-art by 2.36\%.
And after joint optimization, its has another 2.54\% gain.

In scenes of \emph{Camera Jitter}, \emph{Intermittent Object Motion}, \emph{Night Video} and \emph{Thermal}, our method outperforms the others by a large margin. There are lots of camouflaged foreground regions in the cases of \emph{Night Video}, \emph{Intermittent Object Motion} and \emph{Thermal}.
Combining semantic knowledge and motion information, our network can recognize the camouflaged foreground objects.
And the good performance on \emph{Shadow}, \emph{Camera Jitter} and \emph{Night Video} categories demonstrates the effectiveness of our method in dealing with illumination changes, various background changes, and shadows.
Fig.\ \ref{fig:com} shows some visual comparisons of foreground segmentation results.
The segmentation maps of our method have less false detections in the background regions and more refined boundaries in the foreground region.
In the scene of \emph{Shadow}, compared to other methods, our method removes almost all the shadow. For \emph{Bad Weather}, the camouflaged foreground regions are recognized correctly by our method.


\section{Conclusion}\label{sec:conclusion}
We design an end-to-end two-stage deep CNN architecture
for joint background reconstruction and foreground segmentation.
An encoder-decoder network is used to reconstruct the background image, and an MCFCN is introduced to
segment the foreground with the concatenation of the reconstructed background image and the current frame.
At last, the two sub-networks
are jointly optimized with a multi-task loss.
Experimental results on CDnet 2014 demonstrate
the superiority of the proposed approach.

%

\bibliographystyle{IEEEbib}
\setlength\itemsep{-4pt}
\small
\bibliography{zxbib}
\setlength\itemsep{-4pt}
\end{document}